\documentclass[sigconf]{acmart}

\usepackage{xcolor}
\usepackage[normalem]{ulem}

\AtBeginDocument{%
  }

\setcopyright{acmlicensed}
\copyrightyear{2018}
\acmYear{2018}
\acmDOI{XXXXXXX.XXXXXXX}
\acmConference[Conference acronym 'XX]{Make sure to enter the correct
  conference title from your rights confirmation email}{June 03--05,
  2018}{Woodstock, NY}
\acmISBN{978-1-4503-XXXX-X/2018/06}

\begin{document}

\title{Affective Marker Fusion with ASR-Erased Time for Behavioural Ambivalence and Hesitancy Recognition}

\author{Vikas Kumar}
\email{vikas25@iiserb.ac.in}
\orcid{0009-0000-1593-1666}

\affiliation{%
  \institution{IISER Bhopal}
  \city{Bhopal}
  \country{India}
}

\author{Aditya Mishra}
\email{aditya21@iiserb.ac.in}
\orcid{0009-0000-8968-1364}
\affiliation{%
  \institution{IISER Bhopal}
  \city{Bhopal}
  \country{India}
}

\author{Haroon R. Lone}
\email{haroon@iiserb.ac.in}
\orcid{0000-0002-1245-2974}
\affiliation{%
  \institution{IISER Bhopal}
  \city{Bhopal}
  \country{India}
}

\renewcommand{\shortauthors}{Kumar et al.}

\begin{abstract}

Ambivalence and hesitancy (A/H) means being drawn toward and away from a behaviour at the same time. It is a key trigger signal for digital behaviour-change systems. Unlike the canonical emotions studied in affective computing, A/H has no consistent facial signature. It surfaces instead through sparse cues in language, speech, and timing. Prior systems model A/H as cross-modal conflict, pairing several pretrained encoders with explicit disagreement operators. However, none of them isolates whether the conflict operator contributes anything. We run that isolation on the BAH corpus and find it does not help: with encoders and training fixed, changing the operator moves performance less than changing the random seed does, while representation choice matters more and the selection of ensemble weights and decision threshold matters most. Guided by these findings we propose Affective Marker Fusion (AMF), a reliability-gated fusion of affect-specialised representations with interpretable lexical hesitation markers. We also introduce ASR-erased time, a compact representation recovered from ASR chunk timestamps that captures hesitation events the recogniser deletes from the transcript. AMF with AP-weighted ensembling at a fixed threshold reaches 0.731 macro-F1 on the BAH test split, against 0.694 for the existing system. 

\end{abstract}


\maketitle

\section{Introduction}
Behavioural ambivalence and hesitancy (A/H)---the simultaneous pull toward and away from a behaviour---is a central construct in motivational interviewing because it reflects a state of internal conflict in which individuals remain receptive to behavioural change~\cite{rollnick2008motivational, gonzalez2026bah}. Consequently, detecting A/H is an important capability for behaviour-change technologies~\cite{nahum2016just, klasnja2015microrandomized}, including digital coaching applications, remote health interventions, and automated motivational interviewing agents, which rely on recognising when users are most likely to benefit from guidance. Detecting A/H from short video responses is therefore more than an affect-recognition problem; it enables these systems to identify moments when supportive interventions are most likely to be effective. Errors have direct practical consequences, either by missing opportunities for intervention or interrupting users unnecessarily.

Recognising A/H is considerably more challenging than recognising the canonical emotions typically studied in affective computing~\cite{poria2017review, calvo2010affect}. Unlike basic emotions, which often exhibit consistent facial expressions across contexts, A/H has no universal behavioural signature. It is expressed through subtle, temporally evolving cues spanning language, speech, and non-verbal behaviour, including hedged language, self-corrections, prolonged pauses, and qualifying statements. These cues are sparse and unevenly distributed in time, so they are easily diluted when frame- or segment-level representations are pooled into a single video embedding --- a cost that is visible in our results, where every visual representation we evaluate lands in the same narrow band regardless of how it is fused.

We study this problem on BAH~\cite{gonzalez2026bah}, to our knowledge the only public corpus with whole-video A/H annotations, aligned faces, transcripts, and per-segment cue tags. BAH poses the task as binary video-level classification of short interview responses, evaluated by macro-F1 with average precision (AP) as a secondary metric. Its scale is representative of the regime this problem occurs in: A/H annotation requires trained coders and consented health-behaviour interviews, so the 778 labelled training videos are not an artefact of this particular corpus but a structural property of the task. Methods here must therefore generalise from a few hundred examples of a subtle, person-specific behaviour.

Existing approaches largely formulate A/H recognition as a cross-modal conflict problem, hypothesising that ambivalence is reflected by disagreement between what a person says and how they look or sound while speaking~\cite{bekhouche2026conflict,savchenko2025hsemotion}. To capture this, recent BAH systems combine multiple pretrained encoders with explicit conflict operators, typically based on element-wise feature differences~\cite{baltruvsaitis2018multimodal, bekhouche2026conflict,savchenko2025hsemotion}. While intuitive, this design substantially increases model complexity despite the limited size of the available training data. More importantly, although prior work has investigated multimodal fusion and explicit cross-modal conflict for A/H recognition~\cite{gonzalez2026bah,bekhouche2026conflict}, it has not, to our knowledge, isolated the contribution of the conflict operator itself while holding the underlying representations and training procedure fixed.

To answer this question, we systematically reproduce the conflict-modelling pipeline under controlled settings and investigate where the discriminative signal for A/H actually resides. Our study yields three main findings:

\begin{enumerate}
    \item \textbf{Explicit conflict modelling provides limited benefit.} With the encoders and training procedure held fixed, the choice of conflict operator --- absolute difference, orthogonal decomposition, or none --- changes performance by less than retraining one of those configurations under a different random seed, and does not rank consistently across splits.

    \item \textbf{Representation quality matters more than modality combination.} Language is by far the strongest channel; substituting an affect-specialised audio encoder for a generic one is a real gain, while no facial representation we tried lifts the visual channel out of a narrow band close to a prevalence-matched baseline.

    \item \textbf{How the ensemble is selected matters more than how it is built.} Fitting member weights and the decision threshold on the small validation split overfits that split, costing up to 3.4 macro-F1 points on held-out data with the member set held fixed. Avoiding this needs no change to any model.
\end{enumerate}

Together, these findings suggest a different design principle for A/H recognition. Rather than introducing increasingly sophisticated cross-modal interaction modules, it is more effective to first identify where reliable information originates and to calibrate multimodal predictions appropriately. For the BAH challenge, we find that the strongest evidence comes from language and from a previously unexplored source: the temporal artefacts created when automatic speech recognition silently removes hesitations.

Guided by these observations we propose Affective Marker Fusion (AMF), a compact
fusion framework that gates each stream by a learned reliability weight and combines
affect-specialised text, audio, and visual representations with interpretable
hesitation markers. As a separate analysis contribution, we also introduce ASR-erased
time, a temporal signal recovered from ASR timestamps that is the most independent
non-verbal cue we find, although our current fusion does not yet turn that
independence into a score gain.

\begin{itemize}\itemsep2pt

\item \textbf{A controlled empirical analysis} showing that feature representation and calibration contribute substantially more to BAH recognition than explicit cross-modal conflict operators.

\item \textbf{AMF}, a lightweight reliability-gated multimodal fusion framework that combines affect-specialised representations with interpretable hesitation cues while avoiding explicit cross-modal conflict modelling.

\item \textbf{ASR-erased time}, a novel 16-dimensional temporal representation derived from ASR chunk timestamps that captures hesitation information complementary to conventional multimodal features.

\item \textbf{A deterministic training and inference pipeline} achieving \textbf{0.731} macro-F1 on the ABAW 2026 public test set.

\end{itemize}

\section{Related Work}

The ABAW series has progressively extended multimodal affect recognition from frame-level valence-arousal and expression estimation toward more behaviourally grounded targets~\cite{kollias20246th}. The BAH dataset continues this trajectory, replacing categorical emotion labels with whole-video ambivalence/hesitancy annotations, transcripts, aligned faces, and per-segment cue tags~\cite{gonzalez2026bah}, and situates A/H recognition explicitly within digital behaviour-change interventions rather than general affect sensing. The organisers' own baselines -- a zero-shot multimodal LLM ($0.283$ macro-F1) and a co-attention fusion network ($0.590$) -- leave most of the task open, and the systems that have since closed that gap converge on a strikingly similar recipe: three pretrained encoders, a hand-designed cross-modal interaction term, and a validation-tuned ensemble~\cite{bekhouche2026conflict,savchenko2025hsemotion}. Every one of these systems reports a strong aggregate score; none isolates which of its three design choices -- the conflict operator, the encoder selection, or the calibration procedure -- is responsible for it. This paper runs that isolation directly.

\noindent \textbf{Cross-modal conflict fusion:} General-purpose multimodal fusion has long combined modalities under an assumption of complementarity~\cite{baltruvsaitis2018multimodal}, but affect and deception research has argued instead for modelling \emph{disagreement}: multimodal sentiment analysis treats cross-modal conflict as a signal in its own right~\cite{poria2017context}, and deception detection formalises the mismatch between verbal content and non-verbal delivery as a discriminative cue~\cite{perez2015verbal}. The strongest BAH systems import this idea wholesale, computing an element-wise difference (or a related operator) between encoder embeddings as an explicit conflict feature~\cite{bekhouche2026conflict}. Yet the operator itself is never ablated against a no-conflict baseline under otherwise identical encoders and training -- the reported gains could equally come from the extra parameters, the extra branch, or the conflict signal itself. We isolate exactly this by fixing everything except the conflict step (Section~\ref{sec:conflict}), motivated precisely by this gap.

\noindent \textbf{General vs.\ affect-specialised encoders:} Prior BAH solutions build on video and audio backbones pretrained for broad, affect-agnostic objectives -- VideoMAE for generic action recognition~\cite{tong2022videomae} and HuBERT for general speech representation~\cite{hsu2021hubert} -- with the implicit assumption that scale and fusion architecture can recover task-specific signal that the encoder was never trained to expose. Emotion-specialised alternatives exist for audio~\cite{wagner2023dawn} and text~\cite{demszky2020goemotions}, but no BAH system has substituted them in and measured the effect in isolation from architectural changes. We make this substitution directly -- an emotion wav2vec2 encoder for audio, RoBERTa fine-tuned on GoEmotions~\cite{liu2019roberta,demszky2020goemotions} for text, tested against a larger general text encoder~\cite{he2020deberta} -- and find that only one substitution pays off, indicating the field's bottleneck is frequently mis-attributed to fusion when it is really encoder pretraining (Section~\ref{sec:probe}).
\noindent \textbf{Interpretable hesitation cues:} Hesitancy has a well-studied linguistic signature independent of any learned embedding: hedges, filled pauses, self-repairs, and contrastive markers are established correlates of speaker uncertainty in psycholinguistics and computational discourse analysis. Yet every BAH system to date routes language exclusively through a black-box transformer embedding, leaving this readable signal unused and unverifiable at inference time. We compile eleven such markers into a compact, auditable vector that a practitioner can inspect directly, and show it nearly matches a fine-tuned language model on its own (AP $0.800$ vs.\ $0.811$, Table~\ref{tab:streams}) -- evidence that the transformer embedding is not finding much that a short, transparent feature set does not already capture.

\noindent \textbf{Timing artefacts from ASR:} Modern ASR systems such as Whisper~\cite{radford2023robust} are explicitly optimised to emit clean, readable transcripts, and every BAH system that uses transcripts, including ours for the text stream, consumes this cleaned output directly. What this discards is the filler words and hesitation pauses themselves -- precisely the disfluency signal that older, explicitly disfluency-aware speech pipelines were built to detect. No BAH submission recovers this discarded information. We observe that while the deleted \emph{content} is gone, the \emph{time} it occupied is not: it survives as a gap between adjacent chunk timestamps. This motivates ASR-erased time (Section~\ref{sec:erased-method}), a deterministic feature set recovered entirely from timestamps the ASR system already outputs, requiring no additional model or annotation.

\noindent \textbf{Calibration on small data:} Miscalibration is a known failure mode of modern networks even at large scale~\cite{guo2017calibration}, and label smoothing is a standard mitigation~\cite{szegedy2016rethinking}; both issues are magnified when only a few hundred labelled videos are available, as in BAH's $778$-video training split. Prior BAH systems nonetheless follow a common but unexamined ensembling recipe: search member weights and a decision threshold on the validation split, then apply both to test. We show this recipe overfits the $124$-video validation split directly (Section~\ref{sec:calib}) -- a penalty of up to $3.4$ macro-F1 points at a fixed member set, which would go undetected without the controlled comparison we report -- and that removing the threshold search in favour of AP-weighted averaging at a fixed $\tau=0.5$ closes it. Unlike the other two findings, this one requires no new feature or architecture; it is a methodological correction to how prior BAH systems have been evaluating their own ensembles.

\section{Method}

\begin{figure*}[t]
\centering
\includegraphics[width=0.8\linewidth]{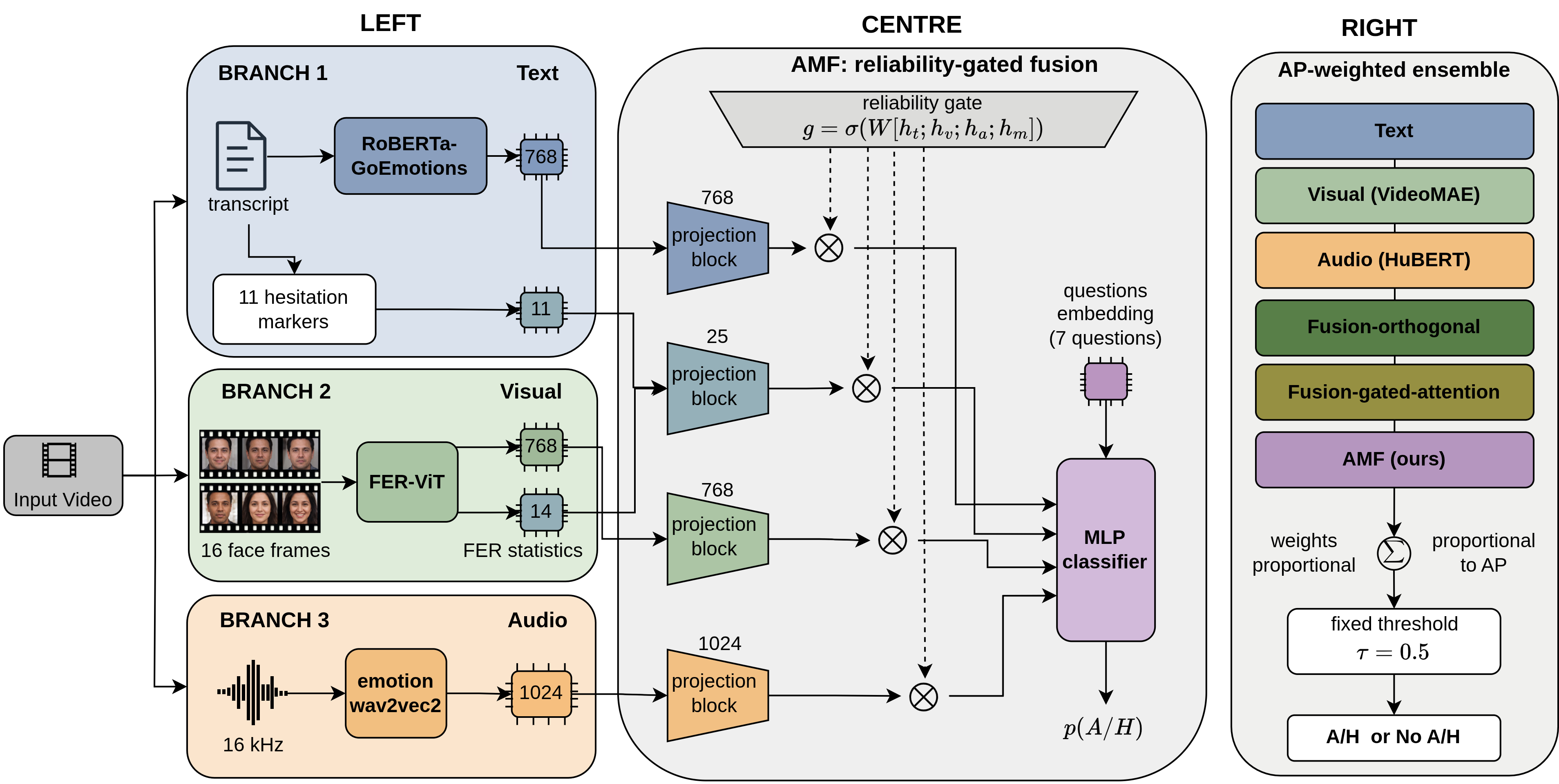}
\caption{Overview. \emph{Left:} frozen affect-specialised encoders. \emph{Centre:} AMF --- per-stream reliability gating, a question embedding, and an auxiliary marker head. \emph{Right:} six-member AP-weighted ensemble at a fixed $\tau=0.5$.}
\label{fig:method}
\end{figure*}

Motivated by the identified gaps, we design our system as separable components that
can each be evaluated on its own. Figure~\ref{fig:method} summarises the resulting
pipeline, and the remainder of this section describes each stage: the frozen
affect-specialised encoders, the reliability-gated fusion mechanism (AMF), the
ASR-erased time feature we introduce, and the calibration-first ensembling scheme
that we show is responsible for most of the final gain.

\subsection{Problem Formulation}
Each video $x$ has a binary label $y\in\{0,1\}$ and an elicitation question
$q\in\{1,\dots,7\}$. We predict a probability $p(y{=}1\mid x)$ and are scored by
$\text{macro-F1}=\tfrac12(F_1^{(0)}+F_1^{(1)})$, with average precision (AP) of the
positive class, the area under its precision--recall curve, as a threshold-free
secondary measure.

\subsection{Modality Encoders}
\label{sec:encoders}

Rather than fine-tuning general-purpose backbones, we give each modality an encoder that was itself trained for affect recognition, and keep all three frozen so that only a small task head has to be learned from the $778$ available training videos (Table~\ref{tab:encoders}). For text, we take the sentence embedding from RoBERTa fine-tuned on GoEmotions. For video, we use a facial-expression ViT~\cite{dosovitskiy2020image}: applied to $16$ aligned face frames per video, it produces both an embedding and a $7$-way expression profile, which we summarise by its mean and standard deviation over time. For audio, we use an emotion-fine-tuned wav2vec2 model~\cite{baevski2020wav2vec,wagner2023dawn}, whose frame-level outputs we mean-pool over time.

For video and audio we use two backbones each. The standalone encoder members and
the two fusion variants use general-purpose backbones (VideoMAE, HuBERT), while AMF
uses affect-specialised ones (FER-ViT, emotion-wav2vec2), so that its errors are less
correlated with the other members' and it adds a more distinct signal to the ensemble.
Text uses a single shared backbone (RoBERTa fine-tuned on GoEmotions) across all
members; AMF additionally consumes an $11$-dimensional hand-crafted lexical
hesitation-marker vector alongside this shared text embedding.

We additionally compute eleven interpretable text markers: rates of hedges,
filled pauses, contrastive (``but''-type) words, negations, first-person
uncertainty, repeated words, and how much sentiment flips within the answer.
AMF's marker stream concatenates these $11$ hesitation markers with the $14$
facial-expression statistics, giving a $25$-dimensional interpretable vector.

\begin{table}[t]
\centering\small
\setlength{\tabcolsep}{10pt}
\caption{Per-stream backbones and feature dimensions. Encoders are frozen and only
lightweight heads are trained}
\label{tab:encoders}
\begin{tabular}{llc}
\toprule
Stream & Backbone & Dim. \\
\midrule
Text        & RoBERTa-GoEmotions~\cite{demszky2020goemotions} & 768 \\
Text markers& hand-crafted hesitation cues                    & 11 \\
Video       & FER-ViT (facial expression)                     & 768 + 14 \\
Audio       & emotion wav2vec2~\cite{wagner2023dawn}          & 1024 \\
\midrule
\multicolumn{3}{l}{\emph{Encoder members (separate) also use:}}\\
Video (enc.)& VideoMAE-Base~\cite{tong2022videomae}           & 768 \\
Audio (enc.)& HuBERT-Large~\cite{hsu2021hubert}               & 1024 \\
\bottomrule
\end{tabular}
\end{table}

\subsection{Affective Marker Fusion (AMF)}
Each stream $s\in\{\text{text},\text{video},\text{audio},\text{markers}\}$ is
projected to a shared width, $h_s=\phi_s(f_s)$. A learned gate assigns each stream a
reliability weight, so weak channels are turned down rather than added as noise:
\begin{equation}
g = \sigma\!\big(W [h_{\text{t}};h_{\text{v}};h_{\text{a}};h_{\text{m}}]\big)
\in (0,1)^4, \qquad \tilde h_s = g_s\, h_s 
\label{eq:gate}
\end{equation}

The gated streams and a question embedding are concatenated and passed to a small MLP. Each question index $q\in\{1,\dots,7\}$ is mapped to a $32$-dimensional vector via a learned embedding table, jointly trained with the rest of the network; it is concatenated with the fused representation immediately before the classification head. An auxiliary head predicts A/H from the $25$-dimensional marker vector alone, which regularises the shared representation and exposes an interpretable sub-decision. 

Training must handle two properties of the data: the two classes are close to balanced in the training split but not in the evaluation splits, and the labels are noisy enough that confident fits are unwarranted at this scale. We therefore minimise a class-balanced binary cross-entropy (BCE) loss with label smoothing over both the fused prediction and the auxiliary marker prediction.

\begin{equation}
\mathcal{L} = \mathrm{BCE}_w\!\left(p,\tilde{y}\right)
            + \lambda\,\mathrm{BCE}_w\!\left(p_{\mathrm{marker}}, y\right)
\label{eq:loss}
\end{equation}
where $\mathrm{BCE}_w(p,y) = -\,\big[\,w\,y\log p + (1-y)\log(1-p)\big]$ is the standard
class-weighted cross-entropy (as in the \texttt{pos\_weight} convention, applying the
weight to the positive term), the positive-class weight $w = N_-/N_+$ is the
negative-to-positive count ratio, and $\tilde{y} = y(1-\epsilon) + \epsilon/2$ is the
smoothed label ($\epsilon=0.1$). We use
the standard smoothing value~\cite{szegedy2016rethinking} rather than tuning it, since $778$
training videos with noisy labels are exactly the regime label smoothing is meant to
guard against overconfidence in. The auxiliary prediction $p_{\mathrm{marker}}$ is obtained by passing the $25$-dimensional interpretable marker vector ($11$ lexical hesitation features $+$ $14$ facial-expression mean/std statistics) through a small projection --- a linear layer followed by layer normalisation, GELU and dropout --- into the shared fusion space, and then through a separate linear classifier trained on that projection alone, so the marker-only decision stays readable. We fix the marker-head weight at $\lambda=0.3$ a priori rather than tuning it (Section~\ref{sec:impl}): with only $778$ training videos we do not have a reliable validation signal for a second loss-weight hyperparameter, so we pick a conventional small value --- well below $1$, so the auxiliary term regularises the shared representation without dominating the main objective --- instead of searching for an optimal one. Note that the main head is trained on the smoothed target $\tilde{y}$ while
the auxiliary marker head is trained on the hard label $y$; this is intentional, so the
interpretable sub-decision stays sharp.

\subsection{Conflict-Fusion Baseline} \label{sec:conflict-baseline} To test the conflict-modelling hypothesis in isolation from AMF, we implement a second fusion architecture that mirrors prior BAH systems~\cite{bekhouche2026conflict}: text, video, and audio embeddings are projected to a shared width and combined via an explicit \emph{conflict step} before an MLP classifier, with no reliability gate and no marker stream. We evaluate three conflict steps under otherwise identical encoders and training: (i) \emph{absolute difference}, the element-wise $|h_i - h_j|$ used in~\cite{bekhouche2026conflict}; (ii) \emph{orthogonal split}, which decomposes each pair of streams into shared and modality-specific components; and (iii) \emph{none}, simple concatenation. We refer to variant (ii) as \textbf{Fusion (orthogonal)} and to a version of this baseline with AMF's reliability gate substituted for the conflict step, but without the marker stream or auxiliary head, as \textbf{Fusion (gated)} -- an ablation isolating the gate's contribution from AMF's other two components. Section~\ref{sec:conflict} reports results.

\subsection{Ensemble and Calibration}
\label{sec:calib-method}
We combine $M$ members by AP-weighted probability averaging and a \emph{fixed}
threshold $\tau=0.5$:
\begin{equation}
\bar p(x) = \sum_{m} w_m\, p_m(x), \qquad
w_m = \frac{\mathrm{AP}_m}{\sum_{m'}\mathrm{AP}_{m'}}.
\label{eq:apw}
\end{equation}
Throughout, by ``calibration'' we mean choosing this operating point, the ensemble
weights and the decision threshold rather than post-hoc probability calibration in
the sense of expected calibration error. AP is threshold-free and therefore stable on
small data. Fixing $\tau=0.5$ is deliberate and principled here: because the
class-balanced loss (Equation~\ref{eq:loss}) re-weights positives to parity, the trained
probabilities are centered so that $0.5$ is the natural, prior-free operating point,
which is partly why the fixed threshold transfers where a validation-tuned one does not
(Section~\ref{sec:calib}). 

\subsection{ASR-Erased Time}
\label{sec:erased-method}

Speech recognisers are trained to output \emph{clean} text: they delete fillers
(``um'', ``uh''), restarts, and hesitation pauses. In BAH, only $7\%$ of the Whisper~\cite{radford2023robust}
transcripts contain any filler word. But the transcripts carry per-chunk timestamps,
and time cannot be deleted: when the recogniser drops a filled pause, the gap it
occupied remains between the end of one chunk and the start of the next. We recover
sixteen deterministic features from these gaps: gap counts and sizes normalised by
speech time, within-chunk speaking rate and its variability, chunk fragmentation,
the gap immediately before a contrast word (the pause-then-qualify pattern), and
restarts across chunk borders. Whisper timestamps reset every $30$\,s, so gaps are
counted only where the timeline is monotone.

These features are computed entirely from the recogniser's own text-side output and use no acoustic signal: no $f_0$, jitter,
shimmer, energy or spectral measurement enters them. They are therefore not prosodic
features in the usual sense, and Section~\ref{sec:probe} shows empirically that they
behave differently from a conventional hand-crafted prosody set.

\subsection{Training on all Labelled Data}
\label{sec:poolfinal}

For the final model we pool train+val+test ($1427$ videos). Since no split remains for early stopping, we carve a seeded stratified holdout of $113$ videos ($8\%$) --- chosen to match the proportion of the original validation split ($124/1427\approx8.7\%$), so the stopping signal is exactly as reliable as before while training sees far more data. We fit the ensemble weights on this holdout and keep $\tau=0.5$, the same fixed, untuned threshold justified in Section~\ref{sec:calib}.

\section{Experiments}

\subsection{Dataset}
BAH~\cite{gonzalez2026bah} contains $1427$ labelled videos from $300$ participants,
split participant-wise (Table~\ref{tab:data}); no person appears in two splits. Every number we report is measured on the
labelled public test ($525$ videos, for models trained on the training split) or on
the shared $113$-video holdout (for the all-data models); we never mix the two.

\begin{table}[!htbp]
\centering \small
\setlength{\tabcolsep}{10pt}
\caption{BAH split statistics (participant-wise).}
\label{tab:data}
\begin{tabular}{lccc}
\toprule
Split & Videos & A/H (+) & No A/H ($-$) \\
\midrule
Train        & 778 & 385 (49\%) & 393 (51\%) \\
Validation   & 124 & 75 (60\%)  & 49 (40\%)  \\
Public test  & 525 & 318 (61\%) & 207 (39\%) \\
\bottomrule
\end{tabular}
\end{table}

\subsection{Implementation Details}
\label{sec:impl}
All models are trained on a single NVIDIA RTX PRO 6000 Blackwell Workstation Edition GPU. The video and audio backbones, and every encoder feeding AMF, are frozen; only lightweight heads are optimized, using AdamW~\cite{loshchilov2017decoupled}, label smoothing~\cite{szegedy2016rethinking} $0.1$, and early stopping (patience $15$). The one
exception is the standalone text member, which also fine-tunes the top four RoBERTa
layers (the lower eight stay frozen) at learning rate $2\!\times\!10^{-5}$ with
R-Drop~\cite{wu2021r}; the
video, audio, and fusion heads use $1\!\times\!10^{-4}$; AMF uses a shared width of
$256$ and marker-head weight $0.3$. We implement everything in
PyTorch~\cite{paszke2019pytorch} with the Transformers library~\cite{wolf2020transformers}. Ensemble weights come from Eq.~\ref{eq:apw} and the
threshold is fixed at $0.5$. Training is deterministic within a fixed
software/hardware state; across sessions individual members can drift by up to about
a point of F1 (GPU numerics), while the ensemble is stable to $\pm0.0001$.
Section~\ref{sec:seeds} retrains the full pipeline under three seeds and quantifies
how much of the variation between our own ablation rows is attributable to the seed
alone.

\subsection{Results}
Table~\ref{tab:main} compares our system with published BAH results. A single AMF
member already exceeds the strongest BAH baseline, and the AP-weighted six-member
ensemble reaches $0.731$ macro-F1 on the public test. A paired bootstrap over the
$525$ test videos gives a $95\%$ interval of $[0.693,0.770]$, and our ensemble
exceeds $0.694$ in $97\%$ of resamples.

\begin{table}[t]
\centering\small
\caption{Macro-F1 on the labelled BAH public test split ($525$ videos). Baseline
scores are as reported in~\cite{gonzalez2026bah, bekhouche2026conflict} on the same split.}
\label{tab:main}
\begin{tabular}{lc}
\toprule
Method & Macro-F1 \\
\midrule
Zero-shot M-LLM (vision only)~\cite{gonzalez2026bah}   & 0.283 \\
LFAN co-attention (V+A+T)~\cite{gonzalez2026bah}        & 0.590 \\
Zero-shot M-LLM + transcript~\cite{gonzalez2026bah}     & 0.634 \\
ConflictAwareAH~\cite{bekhouche2026conflict}            & 0.694 \\
\midrule
Ours: AMF single member                                 & 0.725 \\
\textbf{Ours: 6-member AP-weighted ensemble}            & \textbf{0.731} \\
\bottomrule
\end{tabular}
\end{table}

\begin{table}[t]
\centering\small
\setlength{\tabcolsep}{10pt}
\caption{Individual members on the public test ($525$ videos), each trained on the
$778$-video train split only; the ensemble in Table~\ref{tab:main} combines these six.}
\label{tab:trainonly}
\begin{tabular}{lcc}
\toprule
Member & Test F1 & Test AP \\
\midrule
Text (RoBERTa-GoEmotions)  & 0.721 & 0.873 \\
Video (VideoMAE)           & 0.673 & 0.838 \\
Audio (HuBERT)             & 0.678 & 0.816 \\
Fusion (orthogonal)        & 0.676 & 0.865 \\
Fusion (gated)             & 0.733 & 0.867 \\
AMF (ours)                 & 0.725 & 0.860 \\
\bottomrule
\end{tabular}
\end{table}

\paragraph{Why keep AMF when Fusion (gated) scores higher?}
On this split Fusion (gated) is the best single member ($0.733$ F1 / $0.867$ AP) and
AMF is second ($0.725$ / $0.860$) as shown in Table~\ref{tab:trainonly}. Three observations keep that from being an
argument for dropping AMF. First, the ordering is not one a practitioner could have
acted on: on validation --- the only signal available before test labels exist ---
Fusion (gated) ranks fourth of six ($0.685$), while Fusion (orthogonal) ranks first
($0.693$) and then finishes fifth on test. Picking the member that wins on the test
split is precisely the selection error this paper is about. Second, the gap is not
resolvable at this sample size: a paired bootstrap over the $525$ test videos gives
$\Delta\text{F1}=+0.008$ with a $95\%$ interval of $[-0.031,+0.047]$, with
Fusion (gated) ahead in only $66\%$ of resamples --- a net difference of seven videos. Third, and most
directly, a single-member score is not what either model is for. Adding AMF to the
five-member ensemble improves it under \emph{every} seed we ran
(Figure~\ref{fig:seeds}): $+0.019$ macro-F1 and $+0.007$ AP on average, the AP gain
falling in a tight $[{+}0.006,{+}0.007]$. AMF earns its place as an ensemble
component whose errors differ from the encoder members', not as the strongest member
in isolation.

\subsection{Where is the Signal?}
\label{sec:probe}
We fit a logistic probe on each frozen feature stream, all at a fixed threshold of
$0.5$ (Table~\ref{tab:streams}). Text is strongest (AP $0.811$), and eleven
hesitation markers nearly match it (AP $0.800$). The emotion audio encoder is far
above general HuBERT (AP $0.764$ vs.\ $0.606$; frozen HuBERT collapses to one class
at $\tau=0.5$). ASR-erased time reaches AP $0.718$ --- above every face and
general-audio feature --- and its predictions correlate only $0.11$--$0.36$ with the
six model members, far below anything else we measured. Every visual representation
stays weak, including purpose-built gaze/brow dynamics.

\paragraph{Is ASR-erased time just prosody?}
It is reasonable to ask whether timing features recovered from a transcript are
simply a re-derivation of acoustic prosody. They are not. A conventional
hand-crafted prosody set --- $28$ waveform features covering pause statistics,
articulation-rate variability, $f_0$ and energy variation, jitter, shimmer and
spectral flux --- reaches only AP $0.666$, below the $0.718$ of the $16$ timestamp
features, which touch no audio at all. The two are related but far from
interchangeable: their probe predictions on the test split correlate at Pearson
$r=0.44$ (Spearman $0.45$), whereas the six model members correlate with each other
at $0.84$--$0.97$. What the timestamps recover is the \emph{duration of what the
recogniser deleted}, and that is not something the spectral and voice-quality
statistics of a prosody front-end measure.

\begin{table}[t]
\centering\small
\setlength{\tabcolsep}{10pt}
\caption{Logistic probe on each frozen feature stream (trained on the train split),
at fixed $\tau{=}0.5$, ordered by test AP; feature dimension in brackets. Chance AP
equals the positive prevalence ($0.61$); general HuBERT predicts only the negative
class at this threshold.}
\label{tab:streams}
\begin{tabular}{lccc}
\toprule
Feature stream & Val F1 & Test F1 & Test AP \\
\midrule
Text: RoBERTa-GoEmotions (768)          & 0.632 & 0.646 & \textbf{0.811} \\
Hesitation markers (11)                 & 0.615 & 0.672 & 0.800 \\
Audio: emotion wav2vec2 (1024)          & 0.647 & 0.610 & 0.764 \\
ASR-erased time (16)                    & 0.497 & 0.566 & 0.718 \\
FER statistics (14)                     & 0.529 & 0.551 & 0.669 \\
Audio: hand-crafted prosody (28) & 0.500 & 0.587 & 0.666 \\
Video: AU/gaze (36)                     & 0.538 & 0.540 & 0.665 \\
Video: VideoMAE general (768)           & 0.483 & 0.538 & 0.650 \\
Video: FER embedding (768)              & 0.506 & 0.507 & 0.636 \\
Audio: HuBERT general (1024)            & 0.283 & 0.283 & 0.606 \\
\bottomrule
\end{tabular}
\end{table}

\subsection{Does Conflict Design Help?}
\label{sec:conflict}
We fix the encoders and training and vary only the cross-modal interaction
(Table~\ref{tab:conflict}). The three choices land within $0.05$ AP and do not order
consistently across splits: the orthogonal operator is best on validation but worst on
test F1, while removing conflict entirely gives the best test AP. These gaps are no
larger than the seed-to-seed spread of a single one of these configurations
(Figure~\ref{fig:seeds}: Fusion (orthogonal) alone ranges over $0.050$ macro-F1 across
three seeds), so we read them as ties rather than a ranking.
This does not make the conflict-based fusion members useless: no operator is
individually decisive, yet the two fusion variants remain useful in the ensemble
because their errors are decorrelated from the other members. 

\begin{table}[t]
\centering\small
\setlength{\tabcolsep}{10pt}
\caption{Conflict-operator ablation (public test); same model, only the conflict step
changes.}
\label{tab:conflict}
\begin{tabular}{lccc}
\toprule
Conflict step & Val F1 & Test F1 & Test AP \\
\midrule
Absolute difference~\cite{bekhouche2026conflict} & 0.660 & \textbf{0.717} & 0.828 \\
Orthogonal split                                  & \textbf{0.701} & 0.676 & 0.865 \\
None                                              & 0.683 & 0.712 & \textbf{0.879} \\
\bottomrule
\end{tabular}
\end{table}

\subsection{Calibration Study}
\label{sec:calib}
\begin{figure}[t]
\centering
\includegraphics[width=0.8\linewidth]{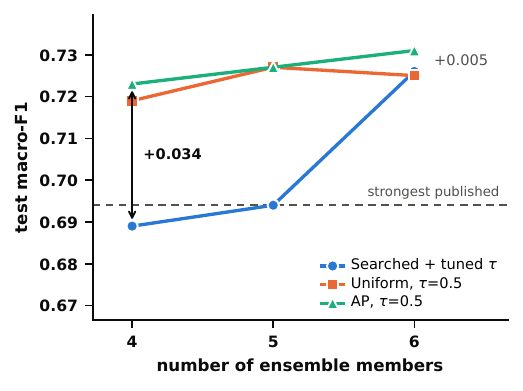}
\caption{Calibration on the public test, with the member set held fixed.}
\label{fig:calib}
\end{figure}

A common ensembling recipe searches member weights and re-tunes the decision
threshold on the validation split, then applies both to the test set --- fitting both
on $124$ videos. Comparing that recipe against ours is only meaningful if the member
set is held fixed, since the two configurations we previously contrasted also
differed in how many members they combined. Figure~\ref{fig:calib} therefore crosses
the two factors, running each recipe on the same $4$-, $5$- and $6$-member pools.

At every member count the searched recipe is the worst of the three, and the ordering
never changes. With the member set fixed, replacing the search with AP weights at
$\tau=0.5$ is worth $+0.034$ macro-F1 on four members, $+0.033$ on five and $+0.005$
on six --- for no change to any model. The gain shrinks as the member count rises because both changes fix the same weakness: with more members, no single member can take most of the weight, so the search cannot go as far wrong. The
failure mode is visible in the weights themselves --- the search puts $71.7\%$ of the
mass on one weak member at four members and $56.1\%$ at six, whereas the AP weights
stay within $0.157$--$0.172$.

The uniform-weight row at each member count separates the weights from the
threshold: it keeps $\tau=0.5$ and only replaces the AP weights with equal ones.
Uniform weights land within $0.006$ of AP weights everywhere, and above the
searched recipe everywhere. The benefit therefore comes from \emph{not fitting the operating point on $124$ videos}, not from the weights themselves. We therefore present Equation~\ref{eq:apw} as a sensible default, not as a contribution. We likewise do not oversell the final margin: a paired
bootstrap~\cite{efron1994introduction} places our $0.731$ above $0.694$ in $97\%$ of resamples, not all of them.


\subsection{Members and Ensemble}
Table~\ref{tab:ensemble} lists the six all-data members on the shared $113$-video
holdout. Each member is a trained head on top of a frozen encoder; the standalone
audio classifier additionally uses learned attention pooling over time, while all
other members and the fusion/AMF inputs use the fixed pooling described in
Section~\ref{sec:encoders}, which is why, e.g., the HuBERT audio member reaches
AP $0.831$ here even though the frozen HuBERT \emph{probe} of Section~\ref{sec:probe}
does not. Adding AMF raises AP from $0.903$ to $0.907$; its effect on holdout macro-F1 is within retraining noise, because the five encoder members are $0.84$--$0.97$ correlated and AMF is still $0.86$--$0.93$ correlated with them. We keep AMF because the leak-free public test favours it under every seed we ran (Figure~\ref{fig:seeds}). The HuBERT audio member and AMF's internal emotion-wav2vec2 front-end play different roles --- a standalone audio member versus a component inside AMF --- and both contribute ensemble diversity.

\begin{table}[t]
\centering\small
\setlength{\tabcolsep}{10pt}
\caption{All-data members on the shared $113$-video holdout; AP weights, fixed
$\tau=0.5$. The holdout serves both early stopping and weighting, so member numbers
are mildly optimistic; the public test (Table~\ref{tab:main}) is the clean evidence.}
\label{tab:ensemble}
\begin{tabular}{lcc}
\toprule
Member & Holdout F1 & Holdout AP \\
\midrule
Text (RoBERTa-GoEmotions)  & 0.788 & 0.906 \\
Video (VideoMAE)           & 0.770 & 0.871 \\
Audio (HuBERT)             & 0.726 & 0.831 \\
Fusion (orthogonal)        & 0.814 & 0.895 \\
Fusion (gated)             & 0.805 & 0.904 \\
AMF (ours)                 & 0.751 & 0.900 \\
\midrule
Ensemble, no AMF (5)       & 0.823 & 0.903 \\
\textbf{Ensemble, with AMF (6)} & 0.814 & \textbf{0.907} \\
\bottomrule
\end{tabular}
\end{table}

\subsection{Seed Sensitivity}
\label{sec:seeds}

\begin{figure}[t]
\centering
\includegraphics[width=0.9\linewidth]{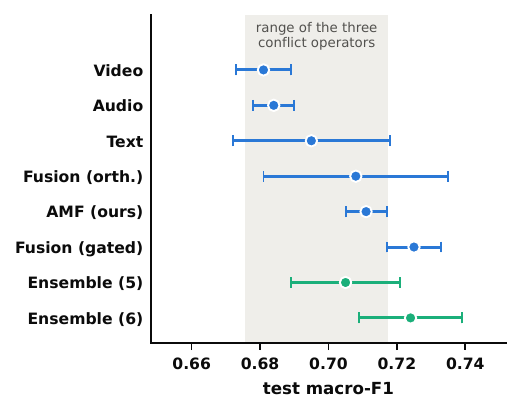}
\caption{Seed sensitivity: mean $\pm$ s.d.\ over seeds $42$, $43$ and $44$.
The band spans the three conflict operators of Table~\ref{tab:conflict}.}
\label{fig:seeds}
\end{figure}

Several of the gaps we report are small, so we retrained the entire train-only
pipeline under three seeds ($42$, $43$, $44$) and give mean $\pm$ standard deviation
on the public test in Figure~\ref{fig:seeds}. The submitted system uses seed $42$
throughout; this is an analysis of its stability, not a change to it.

Three things follow. First, the seed alone moves a single member by as much as
$\pm0.027$ macro-F1 --- Fusion (orthogonal) ranges from $0.677$ to $0.727$ over the
three runs. That spread is wider than the entire conflict-operator ablation of
Table~\ref{tab:conflict} ($0.676$--$0.717$), so that table is measuring seed noise
rather than a property of the operators. This is the conclusion we drew in
Section~\ref{sec:conflict} from indirect evidence, now measured directly, and it is
the strongest support we have for our first finding. Second, ensemble AP is by a
wide margin the most stable quantity in the study ($\pm0.001$, against
$\pm0.003$--$\pm0.019$ for individual members): averaging threshold-free scores is
what buys reproducibility here, while macro-F1 stays more variable because it also
inherits the sensitivity of a single decision threshold. Third, adding AMF improves
the ensemble under every seed in both metrics, and the three-seed ensemble mean
($0.724\pm0.015$) stays above the strongest published system ($0.694$) in every
individual run.

Individual members retrained for this study differ from Table~\ref{tab:trainonly} by
up to $0.008$ macro-F1, the cross-session drift noted in Section~\ref{sec:impl}; the
seed-$42$ ensemble reproduces its $0.731$ exactly. We report the seed-$42$ column in
Tables~\ref{tab:main} and~\ref{tab:trainonly} because that is the single
configuration we fixed in advance and submitted, not because it is the best of the
three.

\subsection{What Did Not Help?}
We list these so others can skip them. Gaze/brow dynamics ($36$ MediaPipe
features~\cite{lugaresi2019mediapipe}) beat the raw FER embedding on a probe but
\emph{lowered} the fused model's public-test AP ($0.860\!\rightarrow\!0.841$).
Prosody features (pauses, jitter, shimmer) pointed the right way but reached only AP
$0.666$ alone and hurt when fused. Supervising an extra head on BAH's cue-type labels
slightly lowered test AP. Swapping encoders gave no gain. Finally, none of the usual
ensemble tricks --- other member subsets, rank or logit averaging, or matching the
predicted positive rate to the training prior ($0.731\!\rightarrow\!0.723$) ---
survived a bootstrap check. ASR-erased time is the one signal we could not yet cash
in: as a seventh member it raises AP ($+0.003$) but not F1, and a stacker fit on the
small validation split gives it a negative weight. We therefore report it as a
diagnostic finding about where unused information lives, not as a component of the
submitted system.

\section{Discussion}

\paragraph{Why is the signal concentrated in text?}
BAH participants answer direct questions about their habits, so when they feel
conflicted they usually say so --- they hedge, qualify, and contradict themselves.
The face and voice carry the same state far more subtly, and $778$ training videos
are not enough to learn subtle cues. Every visual representation we tried, from a
generic video encoder to purpose-built gaze/brow features, lands in the same narrow
band (AP $0.64$--$0.67$), which reads more like a channel ceiling than a modelling
failure.

\paragraph{What we would do differently?}
Set the calibration rules first and touch the model second. Quietly fitting weights
or thresholds on the $124$-video validation split is the most expensive mistake in
this line of work; it cost our own baseline $3.4$ macro-F1 points at a fixed member
set, and it is invisible until you test. We would also invest in ASR-erased time
earlier --- it is the one place where we found information the strong models
demonstrably do not have.

\paragraph{Limitations.}
The public-test margin should be read with its error bars: the bootstrap interval is
$[0.693,0.770]$, so the gain over $0.694$ is probable but not certain. The
$113$-video holdout serves both early stopping and weighting, so its member numbers
are mildly optimistic and small enough that AMF's F1 contribution flips sign between
retrainings there, even though it is positive under every seed on the larger
public test (Figure~\ref{fig:seeds}).

Our evidence also comes from a single corpus. BAH is, to our knowledge, the only
public dataset with video-level A/H annotations, so we cannot currently test whether
these findings transfer to another A/H corpus; what we claim is established for BAH,
and we expect the calibration result to be the most portable of the three, since it
depends only on the size of the validation split and not on anything specific to
this task. Finally, AMF is a small model trained on $778$ videos; nothing here
establishes that reliability gating would still pay off at a scale where the
encoders themselves could be fine-tuned.

\section{Conclusion}

We present a systematic study of behavioural ambivalence and hesitancy recognition on the BAH corpus and draw three conclusions. First, explicit cross-modal conflict modelling provides limited benefit: with the feature encoders and training procedure held fixed, changing the conflict operator moves performance less than retraining the same configuration under a different random seed does, and does not rank consistently across splits. Second, representation quality matters more than modality combination. Affect-specialised audio features provide a meaningful gain, a small set of interpretable linguistic hesitation markers performs competitively with a fine-tuned language model, and none of the facial representations we evaluated lifts the visual channel out of a narrow band. Third, and largest, how the ensemble is selected matters more than how it is built: with the member set held fixed, replacing validation-fitted ensemble weights and decision thresholds with AP-based weighting at a fixed threshold is worth up to $3.4$ macro-F1 points, with no change to any model; a uniform-weight ensemble stays within $0.006$ of it, so the benefit lies in declining to fit the operating point on $124$ videos rather than in the weights themselves. The resulting system, AMF with AP-weighted ensembling, reaches $0.731$ and exceeds the strongest published result on this corpus ($0.694$), though the paired bootstrap interval $[0.693, 0.770]$ places that margin as probable rather than certain.

Beyond these conclusions, our experiments identify \emph{ASR-erased time} as a promising direction: among non-verbal signals it is second only to affect-specialised audio in standalone AP, and by far the least correlated with our other members. We could not yet convert that independence into accuracy on 124 validation videos, but the result suggests that discarded pipeline byproducts are worth revisiting wherever hesitation and uncertainty are the target.

\bibliographystyle{ACM-Reference-Format}
\bibliography{sample-base}

\end{document}